# Ensemble learning with Conformal Predictors: Targeting credible predictions of conversion from Mild Cognitive Impairment to Alzheimer's Disease

Telma Pereira[1,2]
[1]LASIGE, Faculdade de Ciências, Universidade de Lisboa, Portugal
[2]Instituto Superior Técnico, Universidade de Lisboa, Portugal
telma.pereira@tecnico.ulisboa.pt

Sandra Cardoso[3]
[3]Laboratório de Neurociências, IMM and Faculdade de Medicina, Universidade de Lisboa, Portugal
sandradcardoso@gmail.com

Dina Silva[4]
[4]Cognitive Neuroscience Research Group, CBMR, University of Algarve, Faro, Portugal
dlsilva@ualg.pt

Manuela Guerreiro[3]
mmgguerreiro@gmail.com

Alexandre de Mendonça[3]
mendonca@medicina.ulisboa.pt

Sara C. Madeira[1]
sacmadeira@ciencias.ulisboa.pt

## ABSTRACT
Most machine learning classifiers give predictions for new examples accurately, yet without indicating how trustworthy predictions are. In the medical domain, this hampers their integration in decision support systems, which could be useful in the clinical practice. We use a supervised learning approach that combines Ensemble learning with Conformal Predictors to predict conversion from Mild Cognitive Impairment to Alzheimer's Disease. Our goal is to enhance the classification performance (Ensemble learning) and complement each prediction with a measure of credibility (Conformal Predictors). Our results showed the superiority of the proposed approach over a similar ensemble framework with standard classifiers.

## 1. INTRODUCTION
Machine learning is at the core of major advances in the medical and healthcare domain research. In the particular case of Alzheimer's disease (AD), researchers have sought for robust supervised learning models to predict whether a patient with Mild Cognitive Impairment (MCI) is likely to convert to dementia in the future [4, 9]. These prognostic models may then be used to guide clinical decisions in real-world situations concerning patients' treatment, participation in cognitive rehabilitation programs, and selection for clinical trials with novel drugs. Nevertheless, despite the promising results attained by powerful machine learning methods [4, 9], some issues have hampered its practical application in clinical settings. Most methods output the most likely prediction for new examples without providing any assessment of the uncertainty on such predictions. This is a major disadvantage as for clinicians it is paramount to know how much they can trust the prognostic given for a certain patient in order to take decisions. Moreover, these high-standing methods are black-boxes, limiting the clinicians' understanding of the emergent outcome. In this context, researchers should seek for prognostic models which are simultaneously accurate, explainable, and informative regarding predictions' reliability.

Ensemble learning is a commonly used strategy to boost models' performance, drawing much attention in data mining contests [4, 15]. It combines the outcome of multiple classifiers, trained to solve the same problem, in order to make more accurate predictions than individual classifiers. Moreover, Conformal Prediction (CP) is a machine learning technique that complements individual predictions with a measure reflecting how confident the classifier was when classifying such instances [11, 13]. Some studies have evaluated the benefit of combining ensemble learning with conformal prediction, hence harvesting the benefits from both frameworks [2, 6].

In this study, we assess the feasibility of an ensemble-based conformal prediction approach to the prognostic problem of conversion from MCI to dementia, using neuropsychological data. More specifically, we followed a Random Patches (RP)-based ensemble framework [7]. Is it similar to the well-known Bagging method [3] since multiple models, learnt with bootstrap replicates of the data, are generated in parallel. In addition, RP method selects a random subset of features for each classifier. The outcome of these base classifiers is then combined to give a final prediction. Our approach differs from RP method since we use conformal predictors instead of standard classifiers as base estimators. Moreover, we use only credible predictions (above a predefined threshold) to find the final prediction. Each prediction is complemented with a credibility measure which provides insight on its likelihood.

## 2. DATASET AND METHODS
### 2.1 Data
Participants were selected from a revised version of the Cognitive Complaints Cohort (CCC) [12]. This is a prospective study conducted at the Faculty of Medicine of Lisbon to investigate the progression to dementia in subjects with cognitive complaints. Cognitive functioning of participants is evaluated using a neuropsychological battery validated in the Portuguese population, which assesses different cognitive domains, such as memory and reasoning [5]. In total, 41 variables covering clinical, demographic and neuropsychological data were used (described at: https://fenix.tecnico.ulisboa.pt/homepage/ist165127/support-data-for-publications). We used the strategy described in [9] to create learning examples, using time windows. For a given time window, we considered patients that converted from MCI to dementia within a predefined interval, i.e. which had the diagnosis of AD in one of the yearly assessments up until the limit of the window. Those are labeled cMCI (converter MCI). On the other hand, patients that did not convert to AD during that period and presented a diagnosis of MCI at the limit of the window or afterwards, are included in the learning set labelled as sMCI (stable MCI). For a 4-years' time window, 227 (56%) MCI patients remained stable and 175 (44%) converted to dementia.

## 2.2 Conformal Prediction

We introduce the idea behind the conformal prediction framework. For a more formal description we refer to [13]. Let us assume that we are given a training set $\{(x_1, y_1),...,(x_{n-1}, y_{n-1})\}$, where $x_i \in X$ is a vector of attributes and $y_i \in Y$ is the class label (binary classification problem). Given a new test example $(x_n)$ we aim to predict its class. Intuitively, we assign each class $y_n \in Y$ to $x_n$, at a time, and then evaluate how (dis)similar the example $(x_n, y_n)$ is in comparison with the training data. The most likely class label conforms better with the training set under the randomness assumptions. A non-conformity measure must be extracted from the underlying classifier (any machine learning classifier may be used). To evaluate how different $x_n$ is from the training set, we compare its non-conformity score with those of the remaining training examples $x_j$, $j = 1, ..., n - 1$, using the $p$-value function:

$$p(\alpha_n) = \frac{|\{j=1,...,n: \alpha_j \geq \alpha_n\}|}{n} \quad (1)$$

where $\alpha_n$ is the non-conformity score of $x_n$, assuming it is assigned to the class label $y_n$. If the $p$-value is small, then the test example $(x_n, y_n)$ is non-conforming since few examples $(x_i, y_i)$ had a higher non-conformity score when compared with $\alpha_n$. On the other hand, if the $p$-value is large, $x_n$ is very conforming since most of the examples $(x_i, y_i)$ had a higher non-conformity score when compared with $\alpha_n$. Once p-values are computed, CP can be used in one of the following ways:

1) *Using prediction regions.* For a given significance level (ε), CPs output a prediction region, $T^\varepsilon$: set of all classes with $p(\alpha_n) > \varepsilon$. The frequency of errors (fraction of true values outside $T^\varepsilon$) is guaranteed to be less than $\varepsilon$, at a confidence level $1 - \varepsilon$ (under the randomness assumption).

2) *Using forced predictions.* If single predictions are more preferable than prediction regions, CPs predict the class with the highest *p*-value (forced prediction), alongside with its credibility (the largest p-value) and confidence (complement to 1 of the second highest *p*-value). Confidence reveals how likely the predicted classification is compared with the other classes. Credibility reveals how suitable the CP is for classifying the given example. Low credibility means that either the training set is non-random or the test example is not representative of the training set. The probability that the credibility is less than some threshold ε is less than ε (randomness assumption) [11, 13]. The higher the values of both confidence and credibility the more reliable is the prediction.

## 2.3 Ensemble learning

Ensemble learning is based on the assumption that the output emergent from a consensus of methods outperforms that arising from individual methods [15]. Multiple classifiers, named as base estimators, are thus trained to solve the same problem. Then, predictions are obtained by combining the outcome of each base estimator. Two main ensemble paradigms have been proposed to enhance models' performance, either running different base estimators in parallel (such as Bagging and Random Forests) or generating the base estimators sequentially where data distribution is modified, at a time, to increase performance of the base classifiers (such as Boosting) [8, 15].

In parallel-based ensembles, base estimators are fitted on random subsets of the original learning examples and/or of the original feature space, and then each prediction is combined (by majority voting, for instance) to give a final prediction. Such approach reduces variance of base estimators and prevents overfitting. When base estimators are built on both subsets of bootstrapped samples and random features, the method is known as Random Patches (RP) [7]. RP is a simple but successful ensemble method that was originally proposed to handle the problem of insufficient memory regarding the dataset size. It differs from the well-known Random Forests since the subsets of features are selected globally prior to model building while in Random Forests subsets of features are drawn locally at each node of the decision tree classifier. Moreover, any classifier can be used with the Random Patches approach.

## 2.4 Conformal Prediction Ensemble

In this work, we use a supervised learning approach that combines ensemble learning with conformal prediction (Figure 1). By using an ensemble approach we aim to enhance the prediction accuracy w.r.t individual supervised learning models. By exploiting conformal prediction, we aim to: 1) complement each prediction with a measure of credibility and 2) improve the ensemble prediction accuracy by aggregating into the majority voting pool only trustworthy predictions, i.e. base estimators' predictions with credibility above a preassigned threshold. In this context, and assuming that incorrectly classified cases have lower credibility values, we can putatively reduce noise from the classification task. However, as a consequence, some test examples may be considered as unpredictable, if any base estimator is confident enough to classify such example. Our approach is similar to the Random Patches ensemble classifier, in the sense that, both random subsets of learning examples and of features are drawn from the original data to learn the base estimators. This allows users to evaluate the features that are most frequently selected amongst the most trustworthy predictors, for the sake of model interpretability.

Follows a brief description of the proposed ensemble-based Conformal Prediction approach (Figure 1). In the first step, $t$ different subsets of learning examples are drawn with replacement as random subsets of the training data (bootstrapping). Then, a random subset of features is obtained for each subset of examples, $t$. Each dataset is then fed to each base estimator to learn the base models ($M_t$) (train base estimators). In the following step, we combine these models to predict the testing data. Each model ($M_t$) predicts whether a given MCI patient (test example) will convert to dementia (*cMCI*) or remains MCI during the follow-up period (*sMCI*). A measure reflecting the credibility of the predicted class is also outputted. If this value is above a preassigned threshold, the base estimator is considered as trustworthy and thus the predicted class is added to the majority voting pool. Otherwise that prediction is not considered. The final prediction is given by the class that was more frequently assigned by the trustworthy base estimators. Each prediction is complemented with credibility and confidence values, averaged over the experiments. If all predictions are below the predefined threshold, we consider that example as unpredictable and no class is assigned.

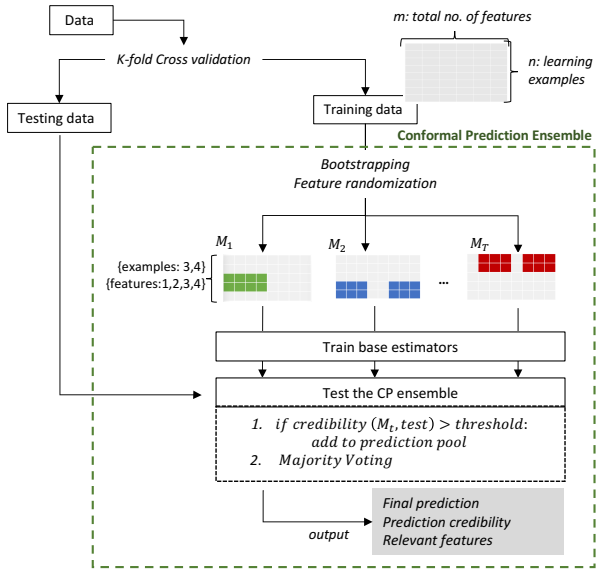

**Figure 1.** Workflow of the ensemble-based Conformal Prediction approach.

### 2.4.1 Classification setup

In this work, a 5-fold stratified cross-validation (CV) procedure repeated for 10 iterations with fold randomization was followed to learn the ensemble-based CP approach. F-measure, sensitivity, and specificity averaged over the 50 iterations are reported as well as the number of unpredictable cases (empty predictions), for each credibility threshold. We used the Naïve Bayes (NB) classifier as base estimator of the ensemble approach since, in previous work [9], it outperformed other commonly used classifiers (such as SVMs, Decision Trees and Random Forests) in the MCI-to-dementia conversion problem. The non-conformity measure is $-\log p(y_i = c|x_i)$, where $p$ is the posterior probability estimated by Naïve Bayes.

In order to assess the potential of the proposed approach, we run the experiments using the following base estimators: 1) conformal predictors using the credibility to reflect uncertainty of each prediction (Forced Predictions), 2) standard classifier with the respective class estimates (posterior probability) to reflect the uncertainty of each prediction, and 3) standard classifiers without considering prediction uncertainty. Moreover, we run the simple classifier for the sake of comparability. Table 1 shows the parameters used in this work. In addition, we bootstrapped 100% of the training data and considered as relevant features selected in 80% of the trustworthy base estimators.

**Table 1.** Parameters used in this study.

| No. of base estimators (CPs) | Randomized features (%) | Confidence level |
|---|---|---|
| {25,50,100} | {25,50,75} | {0.75, 0.80, 0.85, 0.90, 0.95} |

The statistical significance of the classification results was evaluated on the averaged F-measure using the Friedman test to compare results obtained across multiple experiments and the Wilcoxon Signed Rank Test for pairwise comparisons. The classification approach was implemented in Java using WEKA's functionalities.

## 3. RESULTS AND DISCUSSION

Table 2 reports the results of the experimental evaluation obtained with the proposed ensemble approach using CPs and standard classifiers (Naïve Bayes) as base learners with the best parameters found empirically (No. of base estimators: 50, randomized features: 75%). Statistical significant differences were found amongst the experiments ($p < 0.05$, Friedman Test). The CP ensemble outperformed the ensemble using the standard NB, with and without estimates of prediction credibility (posterior probability) ($p < 0.005$, Wilcoxon Signed Rank Test). Naïve Bayes is known to be a simple, yet accurate, classifier. However, NB produces poor probability estimates, often pushed towards 0 and 1 due to the independence assumption [1, 10]. Consequently, posterior probability values cannot properly discriminate the trustworthiness level of predictions. This is corroborated by the small variation of the classification performance over the credibility thresholds. In addition, the percentage of empty predictions is practically zero, since all predictions have high probability estimate values, even if they are incorrectly classified.

Notwithstanding, Naïve Bayes is a good ranker [14]. As aforementioned, non-conformity scores are computed by ranking (and counting the number of) examples in terms of non-conformity to the training data. Hence, despite producing poor class estimators, CPs take advantage of the good ranking abilities of NB, leading to an improvement of the results when using the CP framework. Better results are achieved for higher credibility thresholds (F-Measure: 0.917, sensitivity: 0.927, and specificity: 0.909 for credibility threshold of 0.95). This shows that lower credibility values are given to incorrectly classified classes by CPs. The main weakness of the CP ensemble approach is the large number of unclassifiable cases (empty predictions) which is critical for high credibility thresholds. Being more confident has therefore the cost of having a large number of unclassifiable cases. In the clinical practice, users must find a tradeoff between the quality required on the prognostic prediction and the number of predictions effectively made. NB seems not to benefit much from the ensemble (using random subsets of examples and features) when no credibility threshold is considered, since the results are similar to those obtained with the simple NB.

Table 3 shows an example of the outcome obtained for individual predictions which may be evaluated by clinicians to support their decisions for a given patient. In addition to the credibility and confidence values, the percentage of base estimators that reliably classify that patient can also be seen as a quality measure of such prediction. Furthermore, clinicians may evaluate which features (neuropsychological tests) were more frequently used to predict the type of conversion for that patient.

## 4. CONCLUSIONS

This paper presents a supervised learning approach that combines ensemble learning with conformal prediction. We tested the proposed approach in the prognostic prediction to AD from a stage of MCI, in a real-world dataset. The main purpose was to improve the classification performance of the ensemble by aggregating predictions of only trustworthy base estimators and to complement each individual prediction with a measure of credibility of such prediction. The experimental results demonstrated the superiority of the Conformal Prediction ensemble over the standard ensemble (with and without predictions estimates). Higher credibility thresholds yielded promising results although at the cost of providing prognostic prediction for a small number of patients. We highlight the importance of assigning confidence scores to predictions. Firstly, it is paramount for clinicians to know how much they can rely on the prediction made for a given patient in order to adjust treatments or select patients for clinical trials. Secondly, even unpredictable cases might be useful as clinicians can prescribe more specific exams to deeply evaluate the neurodegeneration of these patients.

## 5. ACKNOWLEDGMENTS

This work was partially supported by FCT under the Neuroclinomics2 project PTDC/EEI-SII/1937/2014, research grant (SFRH/BD/95846/2013) to TP and LASIGE Research Unit ref. UID/CEC/00408/2013.

**Table 2.** Results obtained with the ensemble approach using conformal predictors (values above) and standard (values below) base classifiers, for multiple credibility thresholds. 50 base estimators were learnt with 100% of bootstrapped training set and 75% of the original set of features. Results obtained with ensemble approach with standard Naïve Bayes (NB) with no credibility threshold and simple Naïve Bayes are shown in the 6th and 7th columns, respectively.

| **Credibility** | **0.75** | **0.80** | **0.85** | **0.90** | **0.95** | *Ensemble NB (No credibility)* | *Simple NB* |
|---|---|---|---|---|---|---|---|
| **F-measure** | 0.846±0.01<br>0.803±0.01 | 0.865±0.01<br>0.804±0.01 | 0.886±0.01<br>0.805±0.01 | 0.889±0.02<br>0.806±0.01 | 0.917±0.02<br>0.804±0.01 | 0.805±0.01 | 0.803±0.01 |
| **Specificity** | 0.825±0.02<br>0.838±0.01 | 0.838±0.02<br>0.833±0.01 | 0.869±0.03<br>0.839±0.00 | 0.882±0.02<br>0.841±0.00 | 0.909±0.02<br>0.836±0.01 | 0.837±0.01 | 0.834±0.01 |
| **Sensitivity** | 0.872±0.01<br>0.760±0.02 | 0.897±0.01<br>0.762±0.02 | 0.905±0.01<br>0.761±0.02 | 0.898±0.02<br>0.761±0.02 | 0.927±0.04<br>0.764±0.02 | 0.764±0.02 | 0.764±0.01 |
| **Empty predictions (%)** | 23.31±7.13<br>0.25±0.0 | 34.18±9.09<br>0.27±1.6 | 48.89±1.33<br>0.45±1.3 | 61.08±8.57<br>0.49±0.0 | 75.87±9.23<br>0.52±0.0 | - | - |

**Table 3.** Example of the outcome given for individual predictions obtained with the ensemble approach using CPs ensemble (credibility threshold of 0.85). (%) of trustworthy base estimators represents the number of predictions made above the credibility threshold over the experiments. The reported features are used at least on 80% of the trustworthy base estimators.

| Patient ID | % of trustworthy base estimators | Prediction | Credibility | Confidence | Features |
|---|---|---|---|---|---|
| 1210 | 67% | cMCI | 0.888 | 0.864 | Age, time of complaints, Logical Memory A with Interference (free recall), Word Recall (cued), Orientation, Raven Progressive Matrices |
| 1050 | 47% | sMCI | 0.858 | 0.855 | Gender, Cancelation Task – A's total, Logical Memory Immediate (free recall), Clock Drawing Test, Word Recall (total) |